\documentclass{article}

\usepackage{PRIMEarxiv}

\usepackage[utf8]{inputenc} 
\usepackage[T1]{fontenc}    
\usepackage{hyperref}       
\usepackage{url}            
\usepackage{booktabs}       
\usepackage{amsfonts}       
\usepackage{nicefrac}       
\usepackage{microtype}      
\usepackage{graphicx}       
\graphicspath{{media/}}     
\usepackage{enumitem}        
\usepackage{graphicx}        
\usepackage{caption}         
\usepackage{caption}
\usepackage{subcaption}      
\usepackage{float}           
\usepackage{booktabs}        
\usepackage{multirow, array} 
\usepackage{amsmath, amssymb, amsthm, amsfonts}  
\usepackage{textgreek}       
\usepackage{url, hyperref}   
\usepackage{wrapfig}         
\usepackage{setspace}        
\usepackage{tikz}            
\usepackage{tablefootnote}
\usepackage{etoolbox}
\makeatletter
\patchcmd{\@startsection}
  {\@afterindentfalse}
  {\@afterindenttrue}
  {}{}
\makeatother

\usepackage{titlesec}
\titlespacing{\section}{0pt}{1.5ex plus 0.5ex minus 0.2ex}{1ex plus 0.2ex}

\title{Let’s CONFER: A Dataset for Evaluating Natural Language Inference Models on CONditional InFERence and Presupposition
\thanks{\textit{\underline{Citation}}: This paper is published in the Proceedings of the 38th Canadian Conference on Artificial Intelligence (CAIAC 2025). Please cite the conference version at \url{https://caiac.pubpub.org/pub/keh8ij01}}
}

\author{
  Tara Azin\thanks{Corresponding author: \texttt{taraazin@cmail.carleton.ca}} \\
  Carleton University \\
  \\
  \And
  Daniel Dumitrescu \\
  University of Ottawa \\
  \\
  \And
  Diana Inkpen \\
  University of Ottawa \\
   \\
  \And
  Raj Singh \\
  Carleton University \\
  \\
}

\begin{document}
\maketitle

\begin{abstract}
Natural Language Inference (NLI) is the task of determining whether a sentence pair represents entailment, contradiction, or a neutral relationship. While NLI models perform well on many inference tasks, their ability to handle fine-grained pragmatic inferences, particularly presupposition in conditionals, remains underexplored. In this study, we introduce CONFER, a novel dataset designed to evaluate how NLI models process inference in conditional sentences. We assess the performance of four NLI models, including two pre-trained models, to examine their generalization to conditional reasoning. Additionally, we evaluate Large Language Models (LLMs), including GPT-4o, LLaMA, Gemma, and DeepSeek-R1, in zero-shot and few-shot prompting settings to analyze their ability to infer presuppositions with and without prior context. Our findings indicate that NLI models struggle with presuppositional reasoning in conditionals, and fine-tuning on existing NLI datasets does not necessarily improve their performance.
\end{abstract}


\section{Introduction}
\label{introduction}
Inferring whether a pair of sentences represents an entailment, a contradiction, or a neutral relation is fundamental to Natural Language Inference (NLI). This task has proven to be effective in training and evaluating models on textual reasoning. However, not all forms of inference are handled equally well by NLI models. Among different types of inference, presupposition is a specifically interesting case. Presupposition is a form of pragmatic inference that relies on shared assumptions between speakers rather than purely on logical entailment. It is different from entailment in that it is not canceled when placed in entailment-canceling environments (e.g., negation, questions, modal verbs). Table \ref{tab:presupposition_entailment} shows the distinction between presupposition and entailment under negation, based on standard linguistic theories of presupposition projection \cite{Karttunen1973, Stalnaker1973}. 

Presuppositions become even more complex when embedded in conditional sentences, where the presupposition inherited by the whole sentence appears to be less predictable. Unlike in simple declarative sentences, where presuppositions tend to persist, conditionals introduce dependencies that can sometimes modify the presupposition of an embedded constituent. In particular, sentences in a conditional format like ``\( \text{If } A, \text{then}\ B_p \)'', where p is the presupposition of the consequent (B), appear to presuppose either p itself or ``If A then p'' (further details will be discussed in Section \ref{background}). This variability complicates how presuppositions are interpreted and accommodated in discourse. Understanding how presuppositions behave in such structures is essential for natural language processing tasks, since models must distinguish between context-dependent inferences and inferences that hold unconditionally. 

\begin{table}[h]
\centering
\captionsetup{width=0.75\textwidth}
\begin{tabular}{lcc}
\multicolumn{3}{l}{\textbf{Sentence:} The chocolate cake is delicious.} \\
\multicolumn{3}{l}{\textbf{Negated Sentence:} The chocolate cake is not delicious.} \\
\midrule
 & \textbf{Presupposition} & \textbf{Entailment} \\
\midrule
\textbf{Implied Meaning} & There is a cake. & Something is delicious. \\
\textbf{Remains True After Negation?} & Yes & No \\
\bottomrule
\end{tabular}
\caption{Distinction between presupposition and entailment under negation. The presupposition of a sentence, such as the existence of \textit{a chocolate cake}, remains true even when the sentence is negated. In contrast, the entailment, which follows from the truth condition of the original proposition does not survive negation, and the entailment is canceled.}
\label{tab:presupposition_entailment}
\end{table}

Returning to the definition and function of NLI models, since presupposition involves a specific type of inference that interacts with both semantic and pragmatic reasoning, NLI provides a structured framework to evaluate how well models handle such inferences. It remains unclear whether NLI-trained models truly represent presuppositional reasoning or simply learn surface-level correlations. Therefore, focusing on a specific type of pragmatic reasoning within a fixed linguistic structure could be an effective approach to evaluate a model's ability to handle complex inference tasks. To investigate this, we introduce CONFER, a dataset that is specifically designed to examine machine understanding of presuppositions in certain types of conditional sentences \footnote{\url{https://github.com/Conditional-NLI/CONFER}}. Our dataset consists of 18,000 sentence pairs, generated semi-automatically using linguist-designed templates, allowing us to create a sizable, lexically diverse, and carefully controlled dataset that focuses on specific types of conditional sentences. To our knowledge, no existing NLI dataset is specifically dedicated to conditional sentences. Other available datasets \cite{Jeretic2020IMPpress,Parrish2021} contain either no conditionals or only a limited number, making their evaluation of such structures less comprehensive.

As part of our study, we evaluated two pre-trained models, RoBERTa and DeBERTa, using our data alongside two other presupposition datasets, IMPPRES \cite{Jeretic2020IMPpress} and NOPE \cite{Parrish2021} to assess their performance on pragmatic inference tasks. We also tested four Large Language Models (LLMs): Gemma, Llama, GPT-4o, and DeepSeek-R1, in both zero-shot and few-shot prompting settings to examine their ability to understand specific types of conditionals in contexts with and without prior training. Our findings show that NLI models, including LLMs, struggle to generalize to complex conditional structures, particularly in presuppositional reasoning. While fine-tuning on standard datasets enhances performance on  inference tasks, it does not transfer well to conditionals and often results in performance degradation. Experiments with fine-tuning and prompting techniques indicate that the models have limitations in handling the complexities of conditional reasoning. These results emphasize the need for more diverse and structured datasets to enhance model performance in pragmatic inference.

Our contributions are as follows:
\vspace{2pt}

\begin{itemize}

    \item We introduce \textsc{Confer}, a dataset designed to evaluate pragmatic inference in conditionals.
    \item To the best of our knowledge, this is the first study to evaluate pre-trained NLI models, as well as large language models (LLMs), on presupposition projection in conditional inference.
    \item The dataset, prompts, code, and experimental results are publicly released to facilitate further research in pragmatic and conditional inference within NLI.
\end{itemize}

\newpage
\section{Background}
\label{background}
\noindent\textbf{Pragmatic Inference.} Pragmatic inference extends beyond logical entailment by taking into account contextual and speaker-driven reasoning. It refers to a relationship between two sentences that depends on the context of utterance and the conversational objectives of the speakers. Unlike semantic entailment, which is determined by the logical relationship between sentence meanings in isolation, pragmatic inference results from how language is used in communication and how interlocutors interpret meaning beyond explicit content \cite{Stalnaker1973}.

Among various types of pragmatic inference, presuppositional inference is particularly complex because it interacts with both semantic structure and pragmatic accommodation. A more detailed discussion of presupposition follows below.

\noindent\textbf{Presupposition.} Presupposition is the background assumption that a speaker considers to be true when making an utterance. In semantics, presupposition is
typically viewed as a relationship between two sentences, where the truth of one sentence is a necessary condition for the other to have a truth value \cite{Beaver1997}.

Presuppositions are typically associated with specific linguistic expressions, known as presupposition triggers, which signal their presence in a sentence. These triggers include iterative adverbs (e.g., again, anymore), factive verbs (e.g., realize, know), definite noun phrases (e.g., the king, her book), and change-of-state verbs (e.g., break, crack). For example, in the sentence \textit{Mary failed the exam again}, the iterative adverb \textit{again} signals that Mary has failed the exam before \cite{Potts2015}. Similarly, in \textit{John regrets quitting his job}, the factive verb \textit{regrets} presupposes that John quit his job.

Unlike entailments, presuppositions tend to persist under entailment-canceling environments, such as negation and questions. For instance, \textit{John didn’t realize he was late} still presupposes that John was late. This behavior, known as presupposition projection, distinguishes presuppositions from regular entailment \cite{Karttunen1973}.

Another characteristic of presuppositions is that they can be inferred pragmatically through a process known as accommodation \cite{Lewis1979}. Accommodation occurs when a listener adjusts their background knowledge to accept a presupposed fact as true, even if it was previously unknown. For example, if a speaker says \textit{I need to pick up my brother from the airport}, and the listener was unaware that the speaker has a brother, the listener would accommodate this information as a presupposed fact.

\noindent\textbf{Conditional Presuppositions.} The behavior of presuppositions becomes more complex in certain linguistic structures, such as conditionals. In standard cases, presuppositions tend to project out of embedded clauses, but in conditionals, this projection is less predictable. 

A well-known challenge in this area is the Proviso Problem \cite{Geurts1996} (see \cite{Singh2020} for a recent overview), which arises when a conditional does not seem to inherit the full presupposition of its consequent, but rather a weaker or modified version. For instance, \textit{John will bring his wetsuit} entails that \textit{John has a wetsuit}. However, in a conditional sentence like \textit{If John is a scuba diver, he will bring his wetsuit}, the natural presupposition is that only if John is a scuba diver, then he has a wetsuit. This prediction does not always hold. In a sentence like \textit{If John flies to Toronto, he will bring his wetsuit}, the expected presupposition would be \textit{If John flies to Toronto, he has a wetsuit}. Yet, speakers typically infer that John already owns a wetsuit, regardless of whether he flies to Toronto. This discrepancy explains the Proviso Problem, that questions why presuppositions sometimes project in the expected manner, and other times, listeners accommodate a weaker or modified presupposition based on contextual reasoning. 

How humans interpret presuppositions, particularly in complex structures such as conditionals, remains an open question. The absence of a comprehensive theory complicates the creation of test cases for machine learning models that represent the full diversity of real-world data.

\section{Related Work}
\label{relatedWork}
NLI focuses on determining whether a given hypothesis is entailed, contradicted, or neutral with respect to a premise \cite{Bowman2015SNLI, Williams2018MultiNLI} and has been widely studied as a major task in natural language understanding. A growing number of datasets have been developed to evaluate NLI models’ reasoning abilities. Early datasets such as FraCaS \cite{Cooper1994FraCaS} provided foundational benchmarks, while later large-scale datasets such as SNLI \cite{Bowman2015SNLI} and MultiNLI \cite{Williams2018MultiNLI} expanded the range of sentence types and domains. Other datasets including HellaSWAG \cite{Zellers2018SWAG} and αNLI \cite{Bhagavatula2020aNLI} have aimed to challenge model robustness and reasoning strategies.

Despite these advancements, most NLI datasets focus on classical entailment-based reasoning rather than pragmatic inference, which includes presuppositions and implicatures. IMPPRES \cite{Jeretic2020IMPpress} was the first dataset explicitly designed to examine whether NLI models can recognize pragmatic inferences as entailments. However, its coverage of presuppositions in complex structures such as conditionals is highly limited, as it primarily consists of simple conditional sentences with repetitive consequent. Similarly, NOPE \cite{Parrish2021} explored naturally occurring presuppositions, using sentences extracted from real-world data along with their preceding and following sentences as context. However, its scope did not extend to complex structures like conditional sentences, where presupposition projection becomes more challenging.

Several studies have examined how natural language processing models process presuppositional inferences. Prior research has focused on specific classes of presupposition triggers, such as clause-embedding verbs \cite{White2018Veridicality, Marneffe2019CommitmentBank}, scalar implicatures \cite{Degen2015Implicatures, Schuster2020PragmaticInference}, and adverbial presuppositions \cite{Cianflone2018AdverbialPresuppositions}. While findings suggest that neural models can learn certain presuppositions, they often struggle with context-dependent interpretations, particularly when reasoning beyond surface-level lexical cues \cite{White2018Veridicality, Jiang2019Florence, Marneffe2019CommitmentBank, Ross2019VerbVeridicality}.

In the context of NLI, Jeretič et al. \cite{Jeretic2020BERT} evaluated BERT, InferSent, and Bag-of-Words models on their ability to recognize presuppositions as entailed information. They found that BERT shows some ability to process presuppositions, but NLI models often fail to generalize beyond specific triggers. Parrish et al. \cite{Parrish2021} further demonstrated that context is crucial for presuppositional reasoning, yet existing models struggle to account for the interactions between presupposition triggers and discourse context. None of these studies have addressed presupposition projection in complex syntactic structures, such as conditional sentences.

\section{Data Generation}
\label{data}

Existing NLI datasets do not evaluate presupposition projection in conditionals, despite the significance of this type of pragmatic inference in natural language reasoning. Prior datasets such as IMPPRES \cite{Jeretic2020IMPpress} and NOPE \cite{Parrish2021} have explored presuppositions, but they either focus on general presupposition triggers or use simplified, repetitive conditional structures. For instance, IMPPRES includes conditionals with a fixed structure, \textit{``If [antecedent], that's OK''}, to avoid the complexities of presupposition projection in such sentences. The other corpus, NOPE, which consists of sentences extracted from real-world data, including surrounding context, does not address conditionals at all.

Since no comprehensive NLI dataset exists for studying presupposition projection in conditionals, and given the challenges of extracting such sentences from real-world data, we developed a semi-automatically generated dataset. We followed the approach of Warstadt et al. \cite{warstadt2020blimp} in generating the sentences, except that we used few-shot prompting with OpenAI's GPT-3.5 instead of code-based scripts. In this section, we provide a detailed explanation of the dataset generation process.

\noindent\textbf{Semi-Automatic Sentence Generation.} To generate the dataset, we used OpenAI's GPT-3.5 in a few-shot setting with prompts crafted by two linguists to control sentence structures and lexical properties. A list of 2,139 unique lexical items, annotated with grammatical features, was used to guide sentence formation. The prompt explicitly defined a sentence template for the model. For instance, to generate sentences of Type 3 with possessive trigger, we used \textit{If [subject] has [possession], [subject] will [verb] [PossessiveAdjective] [PossessionType]} to generate a sentence such as \textit{If John has children, he will bring his son to the party}. We then provided the model with two example sentences. This semi-automatic approach helped us control the syntactic and lexical variation in generated sentences.

In total, five types of conditionals were generated based on semantic theories of presupposition projection \cite{Karttunen1973, Stalnaker1973}. The structure of the conditionals followed the form \( S = \text{If } A, B_p \) where \( B_p \) represents a consequent \( B \) containing a presupposition \( p \). The sentence types differ in the logical relationship between A (antecedent) and p (presupposition of the consequent). In designing these structures, we also considered the Proviso Problem, discussed in Section \ref{background}, which arises in conditionals where the relationship between the antecedent and the presupposition of the consequent is not straightforward. Table \ref{tab:conditional_types} provides an overview of these five types with examples.

\begin{table}[h]
\captionsetup{width=0.9\textwidth}
\centering
\begin{tabular}{l p{5cm} p{5cm}}
\toprule
\textbf{Type} & \textbf{Relation between A and p} & \textbf{Example} \\ 
\midrule

\textbf{Type 1} & $A \equiv p$ (A is equivalent to p) & - If Guy has a daughter, he'll bring his daughter to the party. \\
 &  & - If you watched a movie by Nolan before, you'll never watch a movie by Nolan again. \\

\midrule
\textbf{Type 2} & $A \rightarrow p$ (A asymmetrically entails p) & - If the king has a son over 30, the king’s son is bald. \\
 & & - If Tammy listens to this pop song, she'll never listen to a pop song again. \\

\midrule

\textbf{Type 3} & $p \rightarrow A$ (p asymmetrically entails A) & - If John has children, he'll bring his four-year-old daughter to the party. \\
 &  & - If she ever watched a movie, she didn't watch Star Wars again. \\

\midrule
\textbf{Type 4} & A and p are logically independent, \newline but A strongly suggests p & - If John is a scuba diver, he'll bring his wetsuit. \\
 & & - If Mary bought a bird, I wonder how high Mary's bird flies. \\

\midrule

\textbf{Type 5} & A and p are logically and \newline probabilistically independent\tablefootnote{We assume that \( q \) and \( r \) are probabilistically independent if, given a body of knowledge \( K \), the probability of \( r \) does not change by learning \( q \): \(\Pr(q \mid r, K) = \Pr(q \mid K)\).} & - If John flies to Toronto, his sister will pick him up at the airport. \\
 &  & - If Lisa finishes her meeting early, she'll never drive a sports car again. \\

\bottomrule
\end{tabular}
\caption{Five types of conditional sentences with examples. In sentences with possessive triggers (e.g., \textit{his}, \textit{her}, \textit{Mary's bird}) in the consequent, the presupposition (p) refers to ownership (e.g., \textit{the king has a son} in type 2). For sentences with the trigger \textit{again}, p implies a past occurrence of the action (e.g., \textit{she has driven a sports car before} in type 5).}
\label{tab:conditional_types}
\end{table}

For most sentence types, we generated 800 sentences per type, using two presupposition triggers, the iterative adverb \textit{again} and a possessive construction (e.g., \textit{his children}). Each type consists of two sets of 400 sentences, with each set containing one of the triggers in the consequent of the conditional to maintain balance. Due to structural constraints, for Type 4 we generated only 400 sentences using the possessive trigger. In total, 3,600 unique conditional sentences were generated across all types. The generated sentences were then manually reviewed by a linguist for grammaticality and plausibility. Since the list of vocabulary with grammatical features was provided to the model, sentences were generally grammatical and natural. However, the main issue observed was anaphora confusion in some possessive pronouns, which required manual correction.\\[3pt]
\noindent\textbf{Presupposition Projection Embeddings.}
To evaluate presupposition projection, each sentence was embedded in three entailment-canceling environments: negation (\textit{It is not the case that S}), interrogation (\textit{Is it true that S}), and factive embedding (\textit{Sam believes that S}). The embedded sentences were added primarily to test whether the presupposition persists when the conditional is placed within entailment-canceling contexts. Secondly, this approach aligns with other similar datasets \cite{Parrish2021, Jeretic2020IMPpress}.\\[3pt]
\noindent\textbf{NLI Dataset Construction and Labeling.} The dataset was then converted into NLI format by manually writing the hypotheses for each sentence. The hypothesis for each sentence equals \textit{p}, which is the presupposition of the consequent \textit{B}. Each conditional and its embedded forms were assigned the same hypothesis \textit{p}. Additionally, a contradictory hypothesis was included for each unembedded sentence to align with the standard NLI dataset format.  

This process resulted in a dataset of 18,000 sentence pairs\footnote{From the complete dataset, we selected 13,500 pairs to conduct the experiments. This selection was made to ensure a balanced distribution across sentence types and labels.}, to evaluate NLI models and investigate how models handle presupposition projection in conditional sentences. 

Finally, each sentence pair was manually annotated with NLI labels (Entailment, Contradiction, Neutral) by a linguist. A second annotator, a computer science graduate student, double-annotated 20\% of the dataset, and the overall agreement between the two annotators was calculated to be 99.86\%. Table 3 presents sample sentence pairs of different types.

\begin{table}[h]
\captionsetup{width=0.9\textwidth}
\centering
\renewcommand{\arraystretch}{1.2}
\begin{tabular}{p{0.8cm} p{1.3cm} p{6cm} p{3.8cm} p{0.8cm}}
\hline
\textbf{Type} & \textbf{Trigger} & \textbf{Premise} & \textbf{Hypothesis} & \textbf{Label} \\ 
\hline

1 & again & Sam believes that if Nadia adopted a cat, she will never adopt a cat again. & Nadia adopted a cat. & N \\
\hline
2 & possessive & If Marley has an outgoing friend, she travels with her friend. & Marley has a friend. & N \\
\hline

3 & again & If Ryan ever watched a movie, he didn't watch Star Wars again. & Ryan watched Star Wars again. & C \\
\hline
4 & possessive & If George has a dog, I wonder how loud George's dog barks. & George has a dog. & N \\
\hline

5 & possessive & It's not the case that if Scarlett finishes her work early, her brother will take her out for dinner. & Scarlett has a brother. & E \\
\hline
\end{tabular}
\vspace{0.2cm}
\caption{Example sentence pairs from the dataset, showing different conditional types, triggers, unembedded and embedded premises, hypotheses (p), and corresponding label for each pair. Labels are designated as N for Neutral, E for Entailment, and C for Contradiction.}
\label{tab:conditional_examples}
\end{table}

\section{Experiments}
\label{experiments}

To investigate how NLI models and LLMs handle presuppositional inference in conditional sentences, we conducted two experiments. First, we evaluated four different NLI models, including language models, by testing their reasoning capabilities before and after fine-tuning on different datasets, including our data. Second, we performed zero-shot and few-shot prompting with four different LLMs using our dataset to assess their ability to recognize and process presuppositional inferences without explicit training. In this section, we detail the two experiments and discuss their methodologies. The results are presented in Section \ref{results}.

\subsection{Experiment 1: Fine-Tuning NLI Models}

To assess the ability of NLI models in presuppositional inference tasks, we evaluated two baseline models, GloVe \cite{pennington2014glove} and InferSent \cite{conneau-EtAl:2017:EMNLP2017}, as well as two pre-trained transformer models, RoBERTa-large-MNLI \cite{Liu2019RoBERTa} and DeBERTa-large-MNLI \cite{he2021deberta}, which both have been fine-tuned on the MultiNLI corpus \cite{Williams2018MultiNLI}.

GloVe is an unsupervised algorithm for generating word vector representations in which the resulting embeddings represent meaningful linear relationships in the word vector space, and InferSent uses a bidirectional LSTM trained on labeled NLI data. These two models were used to generate sentence embeddings with a neural network consisting of a hidden layer and a classification layer used for classifying the sentence pairs. We trained the baseline models separately on IMPPRES and NOPE and evaluated their performance on our dataset for comparison. The trained models were tested on 20\% of our data (the test set) to assess generalization. Finally, we trained the baseline models on our training set and evaluated their performance on the test set in a controlled setting.

DeBERTa-large-MNLI and RoBERTa-large-MNLI are transformer-based language models fine-tuned on the MultiNLI dataset for NLI tasks, and each contains 355M parameters. We tested these models separately on IMPPRESS, NOPE, and CONFER for comparison purposes. The models were then fine-tuned on the three datasets and tested on CONFER.

\subsection{Experiment 2: Prompting-Based Evaluation}
In this experiment, we evaluated four state-of-the-art language models including Llama-3-8B-Instruct, Gemma-2B-it, GPT-4o, and DeepSeek-R1-Distill-Qwen-1.5B. To assess their ability to handle sentence-level inference and compare their performance with and without prior context, we used both zero-shot and few-shot prompting techniques. The models were tested on the same test set used in Experiment 1. For few-shot prompting, example sentence pairs were drawn from the training set, with models receiving one example per sentence type.

Llama-3-8B-Instruct and Gemma-2B-it are both pretrained and instruction-tuned generative text models that were retrieved from ModelScope and Hugging Face, respectively. GPT-4o, like its predecessors in the GPT-4 series, has been trained with a knowledge base but offers higher processing speed and cost efficiency which makes it a suitable choice for our sentence-level inference task. Finally, we tested DeepSeek-R1-Distill-Qwen-1.5B, a distilled version of DeepSeek-R1, which is a highly discussed model at the time of writing this article due to its reported reasoning capabilities. DeepSeek-R1 adopts a Mixture of Experts (MoE) architecture \cite{Shazeer2017MoE} that enhances computational efficiency and scalability.

To evaluate how these models handle presuppositional inference, we designed a prompt instructing them to assess whether a hypothesis necessarily follows from a given premise, assuming that the premise is true. To avoid priming the models with standard NLI labels, we instructed them to judge the truth value of the hypothesis relative to the premise. The models were asked to output ‘E’ if the hypothesis must always be true, ‘C’ if it must always be false, or ‘N’ if it could be either true or false.

\section{Results \& Discussion}
\label{results}
\subsection{Experiment 1}

Table \ref{tab:non_finetuned_accuracy} represents the results of testing RoBERTa and DeBERTa models on IMPPRES and NOPE, and our dataset prior to fine-tuning. For IMPPRES and NOPE, we used the same test sets as in their original studies \cite{Jeretic2020IMPpress, Parrish2021}, which were retrieved from their respective GitHub repositories. The results indicate that for \textit{Entailments}, model performance mostly remains consistent across all three datasets. However, for \textit{Neutral} and \textit{Contradiction} labels, performance is notably higher on our dataset. The lower results on entailments is likely due to the conditional structure of our sentences, where models show uncertainty in assigning a definite label. Unlike the non-conditional entailment statements on which the models were pretrained, conditionals impose ambiguity in the truth values of their antecedents and consequents which causes models to misclassify neutral examples as entailments, impacting the precision on entailments.

The models were also evaluated on each data type and each trigger (\textit{possessive} and \textit{again}). The results show high performance in predicting the labels for Type 5, where the antecedent and consequent in the conditionals are logically independent. Since the hypothesis does not rely on information from the antecedent, models more frequently classify these pairs as \textit{Entailment}, which aligns with the expected labels for this type. The results of model evaluation, before fine-tuning, at the type and sub-type levels are presented in Table \ref{tab:accuracy}.

\begin{figure}[h]
    \centering
    \includegraphics[width=\textwidth]{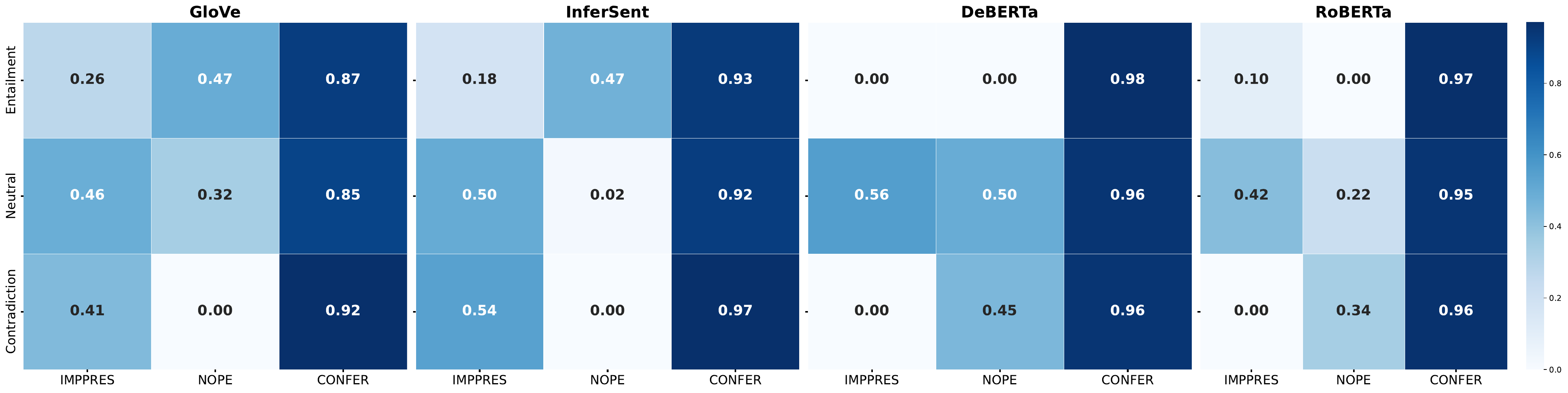}
    \caption{Precision of models trained on IMPPRES, NOPE, and CONFER (tested on CONFER).}
    \label{fig:comparison_chart2}
\end{figure}

The results show significant lower accuracy across all three labels when the baseline models and transformer models are trained on external datasets rather than on our training data. Figure \ref{fig:comparison_chart2} shows the performance of each trained model when tested on our dataset. 

\begin{table}[h]
\captionsetup{width=0.9\textwidth}
\centering
\begin{tabular}{lccc|ccc}
\toprule
& \multicolumn{3}{c}{Tested on IMPPRES and NOPE} & \multicolumn{3}{c}{Tested on CONFER} \\
Model & Entailment & Neutral & Contradiction & Entailment & Neutral & Contradiction \\
\midrule
RoBERTa  & 0.41  & 0.41  & 0.51  & 0.50  & 0.90  & 0.96  \\
DeBERTa  & 0.60  & 0.52  & 0.50  & 0.45  & 0.95  & 0.88  \\
\bottomrule
\end{tabular}
\caption{Precision comparison of RoBERTa and DeBERTa when tested on IMPPRES and NOPE versus our dataset prior to fine-tuning.}
\label{tab:non_finetuned_accuracy}
\end{table}

\begin{figure}[h]
    \centering
    \includegraphics[width=\textwidth]{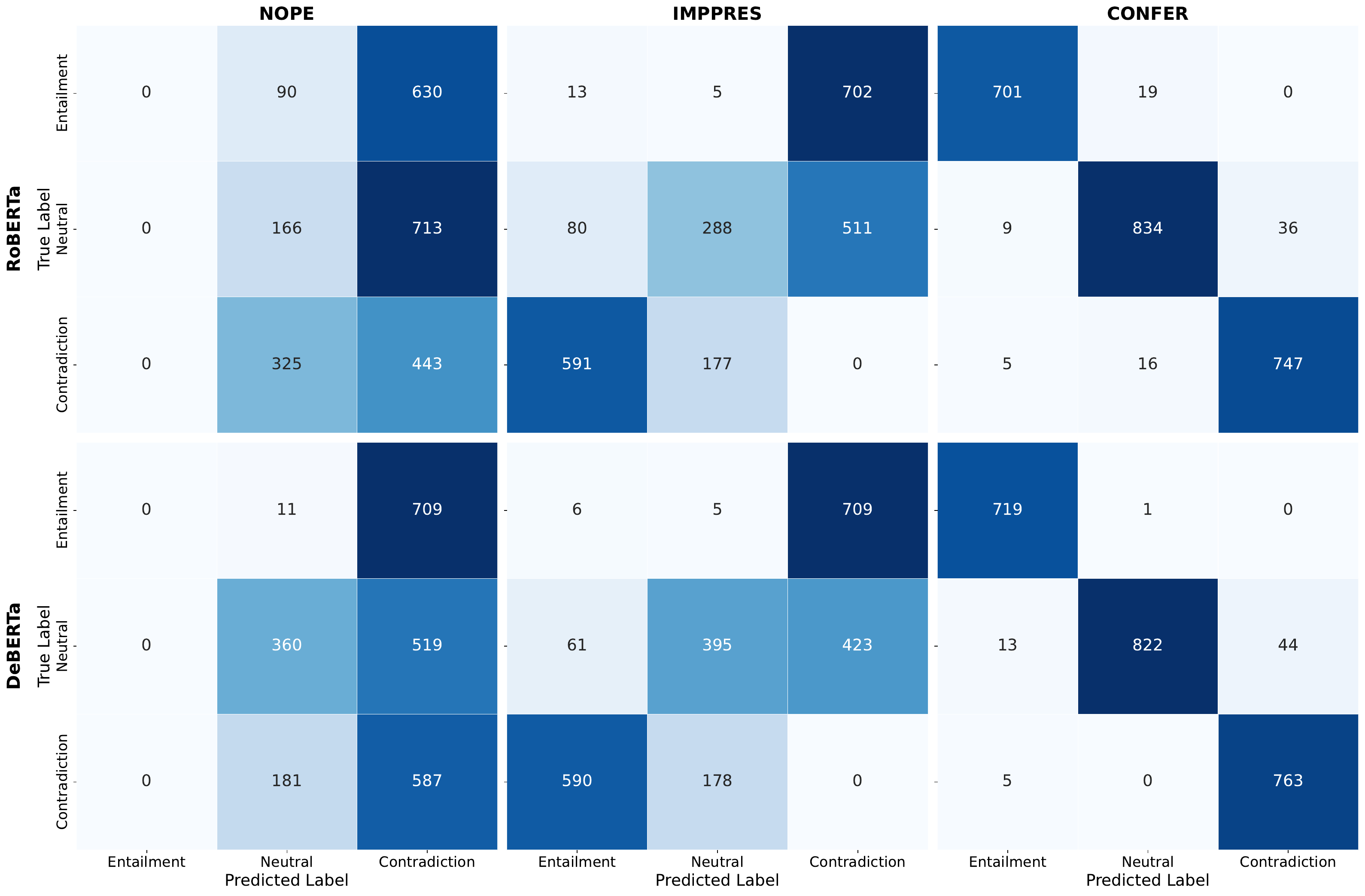}
    \caption{Confusion matrices showing label prediction errors for RoBERTa and DeBERTa fine-tuned on IMPPRES, NOPE, and CONFER.}
    \label{fig:comparison_chart3}
\end{figure}

We closely monitored label prediction errors for each type when fine-tuned on different datasets and tested on CONFER. A high error rate was observed for \textit{Entailment} → \textit{Contradiction} in Type 5. This suggests that models struggle to recognize the entailment relationship when presupposition projection is unclear. It is likely due to the models' reliance on surface-level cues rather than deeper inferential reasoning. Moreover, \textit{Contradiction} → \textit{Neutral} misclassifications were frequent, especially in Type 2, where the antecedent does not necessarily entail the presupposition in the consequent. This pattern implies that models may struggle to classify such cases as \textit{Contradiction} due to the structural similarity with conditionals of other labels. For a detailed breakdown of label prediction errors, see the confusion matrices in Figure \ref{fig:comparison_chart3}.

\begin{table}[h]
\captionsetup{width=0.9\textwidth}
\centering
\begin{tabular}{lcc|cc}
\toprule
& \multicolumn{2}{c}{RoBERTa} & \multicolumn{2}{c}{DeBERTa} \\
Sentence Type & possessive & again & possessive & again \\
\midrule
Type 1 & 0.49 & 0.43 & 0.48 & 0.39 \\
Type 2 & 0.43 & 0.43 & 0.42 & 0.39 \\
Type 3 & 0.47 & 0.43 & 0.41 & 0.40 \\
Type 4 & 0.71 & - & 0.62 & - \\
Type 5 & 0.85 & 1 & 0.84 & 0.97 \\
\bottomrule
\end{tabular}
\caption{Recall of the models across different conditional types and presupposition triggers before fine-tuning. A dash (-) indicates that no items of that type were included in the dataset. See Section~\ref{data} for more details on the Type~4 data.}
\label{tab:accuracy}
\end{table}

\subsection{Experiment 2}
The results of this experiment indicate that while few-shot prompting provides a slight improvement, the models still struggle with inferential reasoning in conditional sentences. This suggests that even state-of-the-art language models are not yet proficient in handling complex inferential tasks compared to human, particularly when reasoning about conditionals. Among all models, GPT-4o achieved the highest accuracy, scoring 64\% in the zero-shot setting and 67\% in the few-shot setting. Table \ref{tab:llm_accuracy} presents the accuracy and F1-scores for each language model.

\begin{table}[h]
\captionsetup{width=0.9\textwidth}
\centering
\begin{tabular}{lcc|cc}
\toprule
& \multicolumn{2}{c}{Zero-shot} & \multicolumn{2}{c}{Few-shot} \\
Model & Accuracy & F1-score & Accuracy & F1-score \\
\midrule
DeepSeek-R1 & 0.32     & 0.32      & 0.36      & 0.36      \\
Gemma      & 0.30  & 0.39  & 0.44  & 0.53  \\
Llama      & 0.52  & 0.61  & 0.39  & 0.54  \\
GPT-4o     & 0.64  & 0.64  & 0.66  & 0.67  \\
\bottomrule
\end{tabular}
\caption{Accuracy and F1-score comparison for zero-shot and few-shot prompting across different LLMs.}
\label{tab:llm_accuracy}
\end{table}

Investigating model performance across different sentence types and triggers shows that all models perform particularly poorly on Type 5 sentence pairs. This is the type for which human annotators assign the \textit{Entailment} label, indicating that the models do not effectively distinguish the Type 1–3 examples (which is frequently identified correctly as \textit{Neutral}) from the Type 5 examples, which should not be \textit{Neutral}. Notably, Llama's performance declines when given examples in the few-shot setting, which suggests that confusion arises from the Type 5 \textit{Entailment} examples provided in the prompt, hindering its ability to classify Type 1–3 examples. The lowest-performing model, DeepSeek-R1, despite being designed for reasoning tasks, fails to generalize well to this inference challenge.

These findings highlight key limitations of language models in inferential reasoning tasks. While LLMs may excel in general reasoning and domain-specific applications, they struggle with fine-grained inference tasks such as those required in NLI. The results further emphasize the importance of creating datasets that focus on linguistic structures and presupposition triggers as a benchmark for evaluating models' understanding of language and logical inference. Tables \ref{tab:model_performance_zero} and \ref{tab:model_performance_few} provide a detailed breakdown of model performance across different sentence types and triggers.

\begin{table}[h]
\centering
\renewcommand{\arraystretch}{1.2} 
\setlength{\tabcolsep}{6pt}
\begin{tabular}{lcccc}
\toprule
\textbf{Type\_Trigger} & \textbf{GPT-4o} & \textbf{Llama} & \textbf{Gemma} & \textbf{DeepSeek-R1} \\
\midrule
type1\_again        & 0.92 & 0.92 & 0.57 & 0.44 \\
type1\_possessive   & 0.66 & 0.75 & 0.30 & 0.41 \\
type2\_again        & 0.88 & 0.92 & 0.54 & 0.45 \\
type2\_possessive   & 0.72 & 0.86 & 0.46 & 0.55 \\
type3\_again        & 0.79 & 0.82 & 0.53 & 0.45 \\
type3\_possessive   & 0.72 & 0.75 & 0.27 & 0.57 \\
type4\_possessive   & 0.52 & 0.41 & 0.26 & 0.31 \\
type5\_again        & 0.35 & 0.03 & 0.07 & 0.11 \\
type5\_possessive   & 0.54 & 0.12 & 0.13 & 0.07 \\
\bottomrule
\end{tabular}
\caption{Recall of the models across different sentence types and presupposition triggers in \textbf{zero-shot} setting.}
\label{tab:model_performance_zero}
\end{table}

\begin{table}[h]
\centering
\renewcommand{\arraystretch}{1.2} 
\setlength{\tabcolsep}{6pt}
\begin{tabular}{lcccc}
\toprule
\textbf{Type\_Trigger} & \textbf{GPT-4o} & \textbf{Llama} & \textbf{Gemma} & \textbf{DeepSeek-R1} \\
\midrule
type1\_again        & 1.00 & 0.60 & 0.62 & 0.56 \\
type1\_possessive   & 0.59 & 0.60 & 0.60 & 0.52 \\
type2\_again        & 0.93 & 0.60 & 0.59 & 0.59 \\
type2\_possessive   & 0.58 & 0.60 & 0.54 & 0.63 \\
type3\_again        & 0.91 & 0.52 & 0.58 & 0.51 \\
type3\_possessive   & 0.56 & 0.60 & 0.60 & 0.59 \\
type4\_possessive   & 0.53 & 0.39 & 0.59 & 0.31 \\
type5\_again        & 0.51 & 0.04 & 0.05 & 0.09 \\
type5\_possessive   & 0.59 & 0.13 & 0.26 & 0.07 \\
\bottomrule
\end{tabular}
\caption{Recall of the models across different sentence types and presupposition triggers in \textbf{few-shot} setting.}
\label{tab:model_performance_few}
\end{table}

\section{Conclusion}
\label{conclusion}
While existing NLI models have been effective for general inference tasks, our results show that both fine-tuned transformers and language models struggle to generalize to more complex syntactic structures, such as conditionals. Although fine-tuning on existing NLI datasets improves performance on similar pragmatic inferences, it does not transfer well to conditionals and often results in decreased accuracy. Our prompting-based evaluation further indicates that even state-of-the-art LLMs fail to reliably infer presuppositional meaning without prior training.

By introducing CONFER, we provide a structured dataset that is specifically designed to assess presuppositional inference in conditionals. The findings from our experiments suggest that improvements in NLI require datasets with more diverse linguistic structures and finer control over inference types. We suggest that future research focus on developing models that can account for the complex interaction between semantics and pragmatics in order to enhance the reasoning capabilities of natural language inference models.

\bibliographystyle{unsrt}  
\bibliography{templateArxiv}  

\end{document}